\documentclass[journal,twoside,web]{ieeecolor}
\usepackage{generic}
\usepackage{cite}
\usepackage{multicol}
\usepackage{multirow}
\usepackage{amsmath,amssymb,amsfonts}
\usepackage{algorithmic}
\usepackage{graphicx}
\usepackage{algorithm,algorithmic}
\usepackage{hyperref}
\usepackage{textcomp}
\usepackage{makecell}
\usepackage{array}
\usepackage{booktabs}
\usepackage{multirow}
\usepackage{relsize}
\usepackage{array, threeparttable, booktabs,  caption}

\begin{document}
\title{OMuSense-23: A Multimodal Dataset for Contactless Breathing Pattern Recognition and Biometric Analysis}
\author{Manuel Lage Cañellas, Le Nguyen, Anirban Mukherjee, Constantino Álvarez Casado, 
        Xiaoting Wu, Praneeth Susarla, Sasan Sharifipour, Dinesh B. Jayagopi, Miguel Bordallo López
\thanks{This work was supported in part by the Research Council of Finland 6G Flagship program under Grant 346208, and in part by PROFI5 HiDyn program under Grant 326291.}
\thanks{Manuel Lage Cañellas, Le Nguyen, Constantino Álvarez Casado, Xiaoting Wu, Praneeth Susarla, Sasan Sharifipour and Miguel Bordallo López are with the Center for Machine Vision and Signal Analysis (CMVS), Infotech Oulu, University of Oulu, Finland. Anirban Mukherjee and Dinesh B. Jayagopi are with Multimodal Perception Lab, IIIT Bangalore, India.}}
\maketitle

\begin{abstract}
In the domain of non-contact biometrics and human activity recognition, the lack of a versatile, multimodal dataset poses a significant bottleneck. 
To address this, we introduce the Oulu Multi Sensing (OMuSense-23) dataset that includes biosignals obtained from a mmWave radar, and an RGB-D camera. The dataset features data from 50 individuals in three distinct poses  -standing, sitting, and lying down- each featuring four specific breathing pattern activities: regular breathing, reading, guided breathing, and apnea, encompassing both typical situations (e.g., sitting with normal breathing) and critical conditions (e.g., lying down without breathing). In our work, we present a detailed overview of the OMuSense-23 dataset, detailing the data acquisition protocol, describing the process for each participant. In addition, we provide, a baseline evaluation of several data analysis tasks related to biometrics, breathing pattern recognition and pose identification. Our results achieve a pose identification accuracy of 87\% and breathing pattern activity recognition of 83\% using features extracted from biosignals. The OMuSense-23 dataset is publicly available as resource for other researchers and practitioners in the field.
\end{abstract}


\section{Introduction}
\label{sec:introduction}

The development of human sensing technologies has become increasingly relevant for monitoring individuals, particularly those who live alone \cite{gokalp2013monitoring}.
The use of wearables for biosignal monitoring has gained popularity due to the availability of reliable sensing infrastructures \cite{wang2019deep} and their facilitation of comfortable monitoring in a continuous manner \cite{hwang2016feasibility}.
However, the use of wearables in biosignal acquisition for human monitoring poses several challenges. They need to be worn continuously, which can lead to monitoring gaps when the battery needs charging or the device is left in an inaccessible place. 

Currently, there are no wearables for breathing pattern analysis specifically designed to be worn for long-duration periods of time. Breathing monitoring typically requires the use of large devices intended for short-term applications, rendering them impractical for continuous monitoring \cite{da2019breathing}. This leads to unbalanced volume of data between biosignal and breathing patterns when monitoring an individual. Non-contact technologies offers a practical alternative that enables continuous monitoring for biosignals \cite{nguyen2023non} and breathing patterns, avoiding the drawbacks associated with wearables, such as potential discomfort.

Visual modalities, including RGB \cite{aggarwal2014human} and depth cameras \cite{kempfle2021breathing}, provide rich appearance information for biometric signs such as pulse and breathing pattern estimation, albeit limited by viewpoints, illumination conditions \cite{szankin2018long}, and privacy concerns.
RGB cameras can be used to extract remote photoplethysmography (rPPG) and remote ballistocardiography (rBCG) signals from the captured video \cite{casado2023depression} allowing for the monitoring of vital signs without the need to transmit or store the video itself.

Radar technology complements the data captured by visual sensors by offering a reliable means to monitor human activities and vital signs \cite{fernandes2022survey} in environments where visual methods may be less effective, such as low-light condition or obstructed space.
Its use is particularly advantageous for privacy since radars capture physiological changes without producing detailed images of the individual.
Integrating visual and radar sensors with signal processing and machine learning algorithms significantly enhances the accuracy and reliability of physiological signals and vital sign monitoring \cite{haque2020illuminating}.

Despite the benefits of non-contact biometrics and breathing analysis technologies, their development and adoption face a major challenge: the lack of extensive, multimodal datasets specifically designed for breathing activity recognition, particularly those that encompass emergency scenarios, for example individuals experiencing apnea while lying down. These datasets are essential for improving algorithms and models to accurately analyze the complex data gathered through non-contact methods. Furthermore, the absence of gender-balanced datasets may contribute to the generation of biased machine learning algorithms due to the underrepresentation of women \cite{leavy2018gender}.

To address this gap, we present the Oulu Multi Sensing Lab Dataset (OMuSense-23), a novel multimodal dataset explicitly designed for non-contact biometrics and breathing pattern analysis. The key contributions of our work are threefold: 

\begin{itemize}
\setlength\itemsep{0pt}
\setlength\parskip{0pt}

\item A novel dataset comprising biosignals obtained from an RGB-D camera and a millimeter Wave (mmWave) radar data collected from 50 participants. The data capture process involves 50 subjects with a balanced sex distribution (50\% male, 50\% female) and a varied range of ages comprised from 24 to 65 years. The subjects were engaged in four breathing pattern activities (normal breathing, reading, guided breathing, and breath holding to simulate apnea) each one performed in three distinct static poses: standing, sitting, and lying down. We describe in detail our data collection protocol to support the replication of the dataset in alternative locations. 

\item A proposal of different statistical and physiological features from remote biosignals. We include well-established statistical measures such as mean, median, and standard deviation (SD), along with advanced features such as fractal dimensions and entropy measures to capture the complexity of biosignals. Additionally, we include biosignal-specific features based on heart rate and respiratory variability extracted from cardiac and breath waveforms.

\item A baseline benchmark evaluation for multiple tasks, including pose identification, breathing pattern recognition, and regression of biometric and physiological characteristics, such as height, weight and age. The evaluation includes detailed training and testing protocols. 

\end{itemize}

To the best of our knowledge, this is the first multimodal dataset capturing biometric characteristics and breathing pattern analysis, encompassing both typical situations (e.g., sitting with normal respiration) and critical conditions (e.g., lying down without breathing). The dataset is made publically available in Zenodo \cite{lage2023omusense} (https://zenodo.org/records/11115819).

\section{Related Work}
\label{sec:relatedwork}

Existing human monitoring datasets typically emphasize two primary tasks: vital signs measurement or activity recognition \cite{nguyen2023non}.
Vital signs measurement is often addressed through unimodal datasets that utilize either cameras \cite {selvaraju2022continuous} or radars \cite{soto2022survey} separately, with limited focus on individual physiological parameters such as height or weight. Extensive research has been dedicated to activity recognition using computer vision \cite{robertson2006general} or radar waves \cite{li2019survey}. While these efforts often prioritize the detection of human gestures significantly more than static poses at home, they even less frequently explore poses in conjunction with different breathing patterns.

Studies focusing solely on respiratory patterns have utilized radar and camera modalities. For example, in \cite{batra2022respiratory}, mmWave radar has been employed to estimate respiratory patterns, with a specific focus on abnormal breathing patterns such as tachypnea and bradypnea. Similarly, RGB cameras have been utilized for breathing pattern recognition, as demonstrated in \cite{islam2023respiratory}, where various abnormal respiratory patterns were accurately recognized. Nevertheless, studies on breathing patterns tend to prioritize the investigation of abnormal respiratory behaviors indicative of illness or chronic conditions, rather than focusing on the study of variations of normal breathing patterns observed in individuals during critical situations.

Emergency poses, particularly lying down positions, have also been explored using RGB cameras, as seen in~\cite{Kepski2014}. However, existing datasets for measuring vital signs, including respiratory and cardiac data, are often constrained in terms of subject poses. For example, datasets such as FCS21~\cite{Kempfle2021}, UBFC-rPPG~\cite{ubfc}, UBFC-Phys~\cite{phys}, and UCLA-rPPG~\cite{ucla} typically capture data when subjects are either sitting or standing, limiting their applicability to emergency situations.

Recent advancements have seen the emergence of multimodal datasets that combine both non-contact and contact sensing modalities. For example, UTD-MHAD \cite{Chen2015} and MHAD \cite{Ofli2013} have employed Kinect cameras and wearable sensors for activity recognition. However, there is now a growing interest in datasets that rely solely on non-contact sensing modalities, as this approach reduces the need for individuals to carry sensors.

OPERAnet~\cite{Bocus2022} represents this trend, utilizing a combination of WiFi sensing devices, ultra-wideband impulse radar, and Kinect motion sensors for activity recognition. Despite its diverse modalities, it lacks information on breathing patterns estimation and involves only six participants, indicating a need for datasets that encompass a wider range of physiological parameters, including breathing patterns involving larger participant groups.

Another common issue of the existing datasets is that the ratio of female and male subjects is not balanced. For example, Kempfle and Laerhoven~\cite{Kempfle2021} provided a dataset of 12 male and seven female participants, while CZU-MHAD~\cite{Chao2022} only has data of male participants.

To address these limitations, our proposed dataset captures breathing patterns in different positions, including sitting, standing, and lying down, simulating emergency situations like apnea while a subject is lying down. We utilize mmWave radar and RGB-D camera to collect data for biometric analysis and breathing pattern recognition while addressing the sex ratio issue by recruiting an equal number of female and male participants.

\section{Data Collection}
\label{sec:databasecollection}

This section explores the details of our recruitment procedures, the sensors used and the methodologies employed for defining the activities recorded for each participant, including sensor placement and the protocols for recording the participants activities.
 
The data collection process involved the replication of monitoring scenarios observed in domestic settings. Breathing patterns and biometric analysis are highly susceptible to motion interference, posing significant challenges for analysis during movement. As a result, to minimize noise and focus on chest displacement induced by respiration and heart activity, participants in our dataset are captured in stationary positions. The combination of poses and breathing patterns thus encompasses regular scenarios such normal breathing or reading while sitting down and emergency scenarios, such as individuals experiencing apnea while lying down. Additionally the collection of biometric characteristics allows the estimation and stratification of users based on their biometric traits.

\subsection{Participants and recruitment process}

A total of 50 participants were recruited using the word of mouth method, as described by Manohar et al. \cite{manohar2018recruitment}. Each participant filled themselves a form with age, sex, height, weight and country of origin. In addition, the authors measured chest circumference, shoulder distance, temperature and blood pressure using a common tape, an infrared thermometer and a portable blood pressure monitor device. We ensured a 50\% balanced sex distribution, with 25 males and 25 females from 14 different countries: China (12), Finland (9), Spain (7), India (7), Vietnam (3), Iran (3), Bangladesh (1), Brazil (1), Bulgaria (1), Canada (1), Ecuador (1), Italy (1), Pakistan (1), Russia (1) and Ukraine (1). The age range of the participants is between 21 and 65 years. Table \ref{tab:demographics} provides a detail of their demographics. 

Ethical considerations were strictly followed in conducting the data collection,  according with the Declaration of Helsinki concerning aspects of research ethics and informed consent and adhering to the regulations set forth by our institutions. The measurements were distributed throughout all working hours, spanning from 8:00 to 18:00. Prior to their participation, all individuals were fully informed about the objectives of the research and provided their informed agreement by signing a consent form. No compensation was provided for their involvement in the study.

\begin{table}[ht!]
\begin{center}
\renewcommand{\arraystretch}{1.2}
\setlength{\tabcolsep}{1.0em}
\caption{Mean and [SD] of the dataset demographics.}
\begin{tabular}{l|l|l|l}
 & All & Male & Female\\
\hline
\# Participants        &50 &25 &25 \\
Age                 &31.04 [7.04]   &31.20 [8.53]   &30.88 [5.33]\\
Height (cm)         &169.20 [9.55]  &174.64 [9.34]  &163.76 [6.11] \\
Weight (kg)         &68.24 [14.00]  &74.80 [12.51]  &61.68 [12.42] \\
\end{tabular} 
\label{tab:demographics}
\end{center}
\vspace{-5mm}
\end{table}

\subsection{Measuring devices}\label{Sensors}

The data acquisition process encompasses the utilization of a set of two non-contact sensors: an RGB-D camera and an mmWave radar. Both devices are located on an external table pointing at the face and chest of the user. The RGB-D camera is utilized to capture facial and chest images in dual RGB and depth mode, while the radar system, in its vital signs configuration, is employed to capture breathing and heart waveforms as well as the chest displacement of the participant.
Both sensors are synchronized via software. 

\begin{figure}[h]
  \begin{center}
    \includegraphics*[width=0.5\columnwidth]{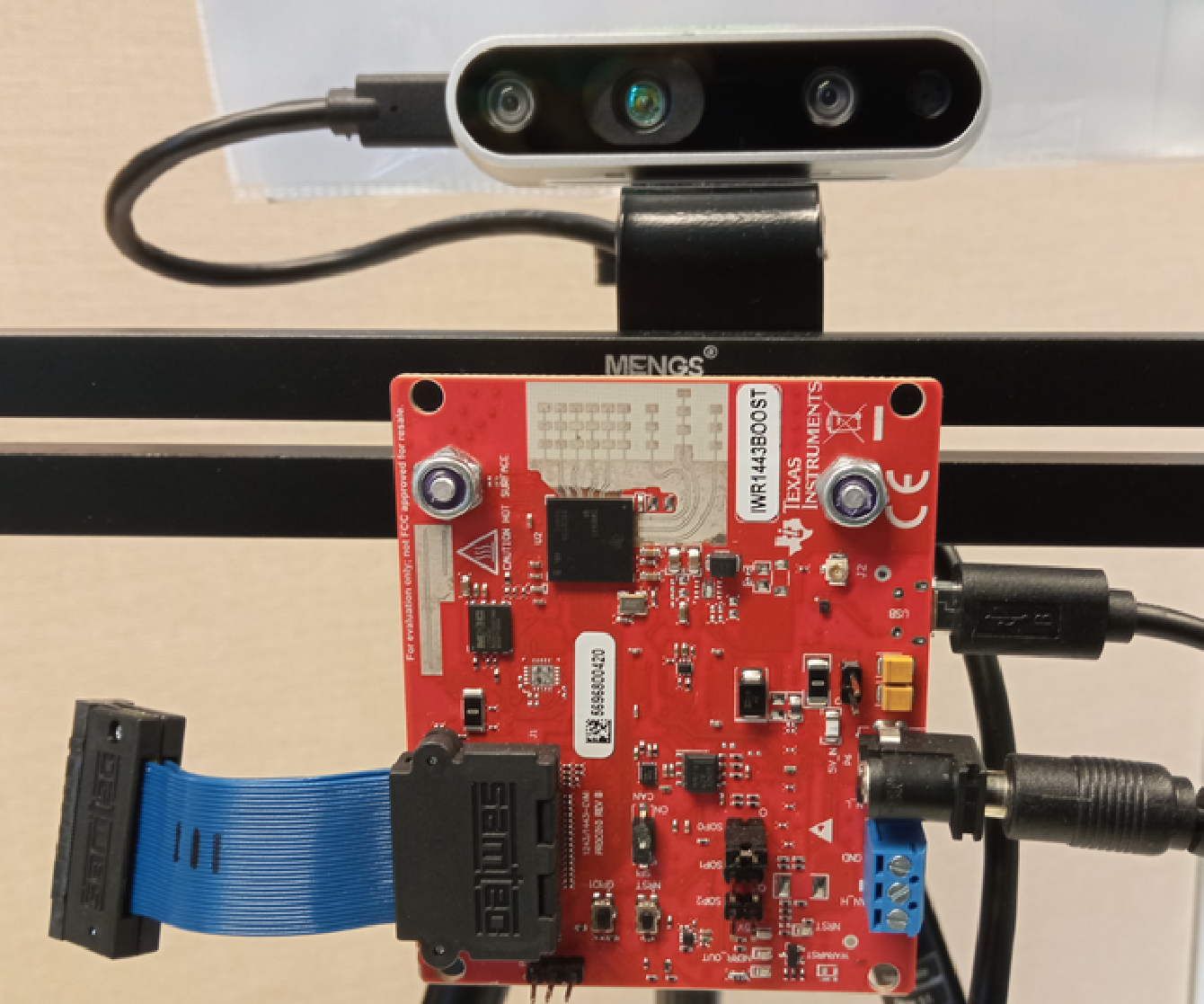}
  \end{center}
  \vspace{-3mm}
  \caption{System setup containing an RGB-D Camera Intel Realsense D435 and an mmWave Radar Texas Instruments IWR1443 in a height configurable tripod.}
  \label{fig:tripod}
  \vspace{-3mm}
\end{figure}

The remote measurement devices are mounted in the same tripod with manual configurable height that is placed in front of the user as shown in Fig. \ref{fig:tripod}. The tripod is placed in a height adjustable standing desk. 

\subsubsection{RGB-D camera}
\begin{figure}[h]
  \begin{center}
    \includegraphics*[width=0.65\columnwidth]{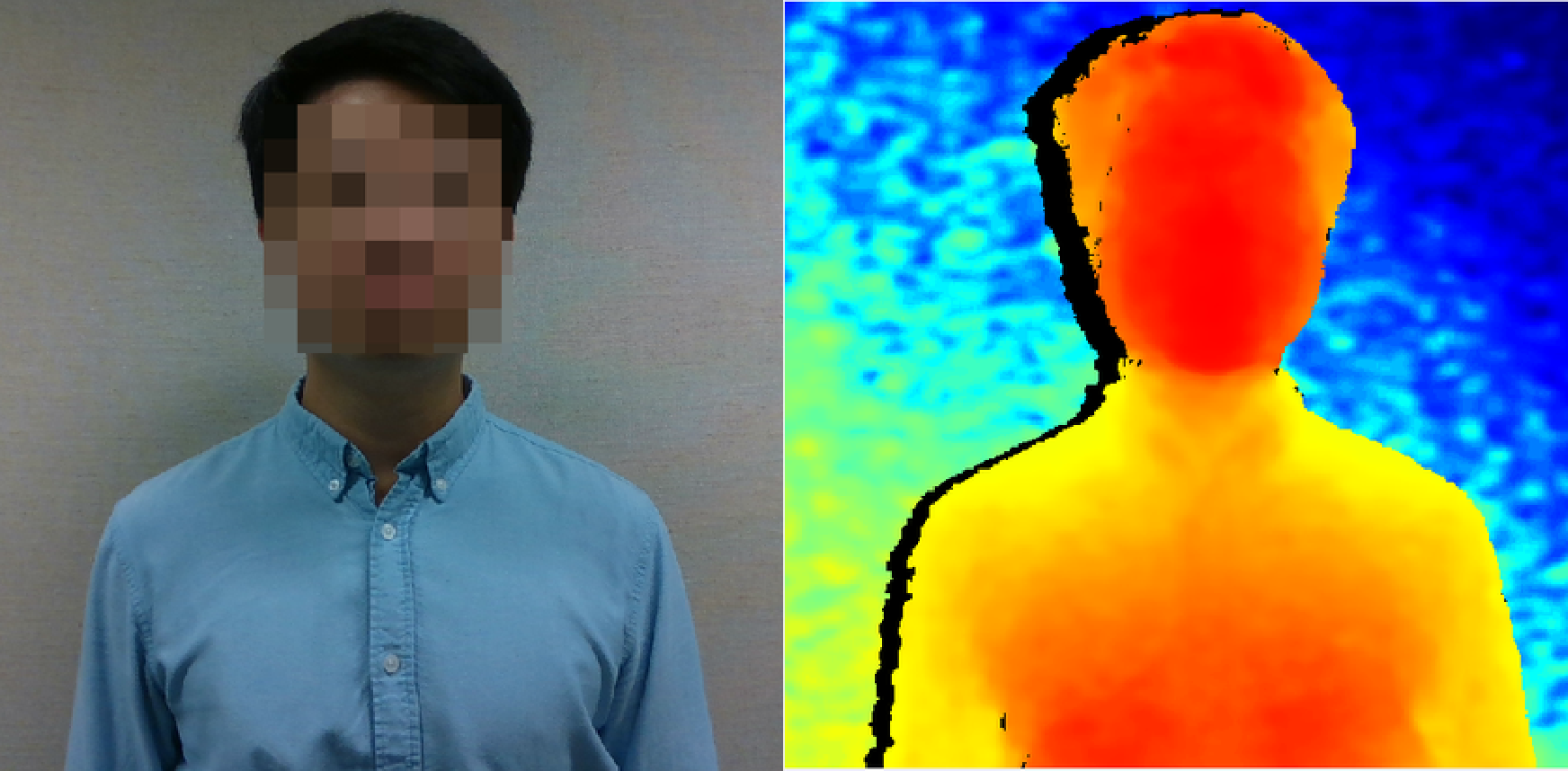}
  \end{center}
  \vspace{-3mm}
  \caption{RGB-D Camera stream: RGB data (left) and depth information (right).}
  \label{fig:rgb_d}
\end{figure}

Intel RealSense D435 is an RGB-D camera designed for capturing both color and depth information. The RGB sensor captures images with a resolution of 1280 x 720 pixels, operating at a rate of 30 frames per second (FPS). The depth images are calculated using active stereo vision, where an infrared (IR) projector emits a pattern of dots onto the scene. Two infrared sensors capture the reflected IR light to calculate the disparity in the images providing an image of 1280 x 720 pixels resolution at a frame rate of 30 FPS, synchronized with the RGB images. Depth information is coded in 8-bit pixels mapped in a particular colormap (JET). We compress each image data stream into two separated \textit{mp4} video files. Concurrently, the timestamps of each frame are stored in a separate \textit{csv} file. An example of the two streams provided by the camera can be shown in Fig. \ref{fig:rgb_d}.

\subsubsection{mmWave Radar}
\begin{figure}[h]
  \begin{center}
    \includegraphics*[width=1\columnwidth]{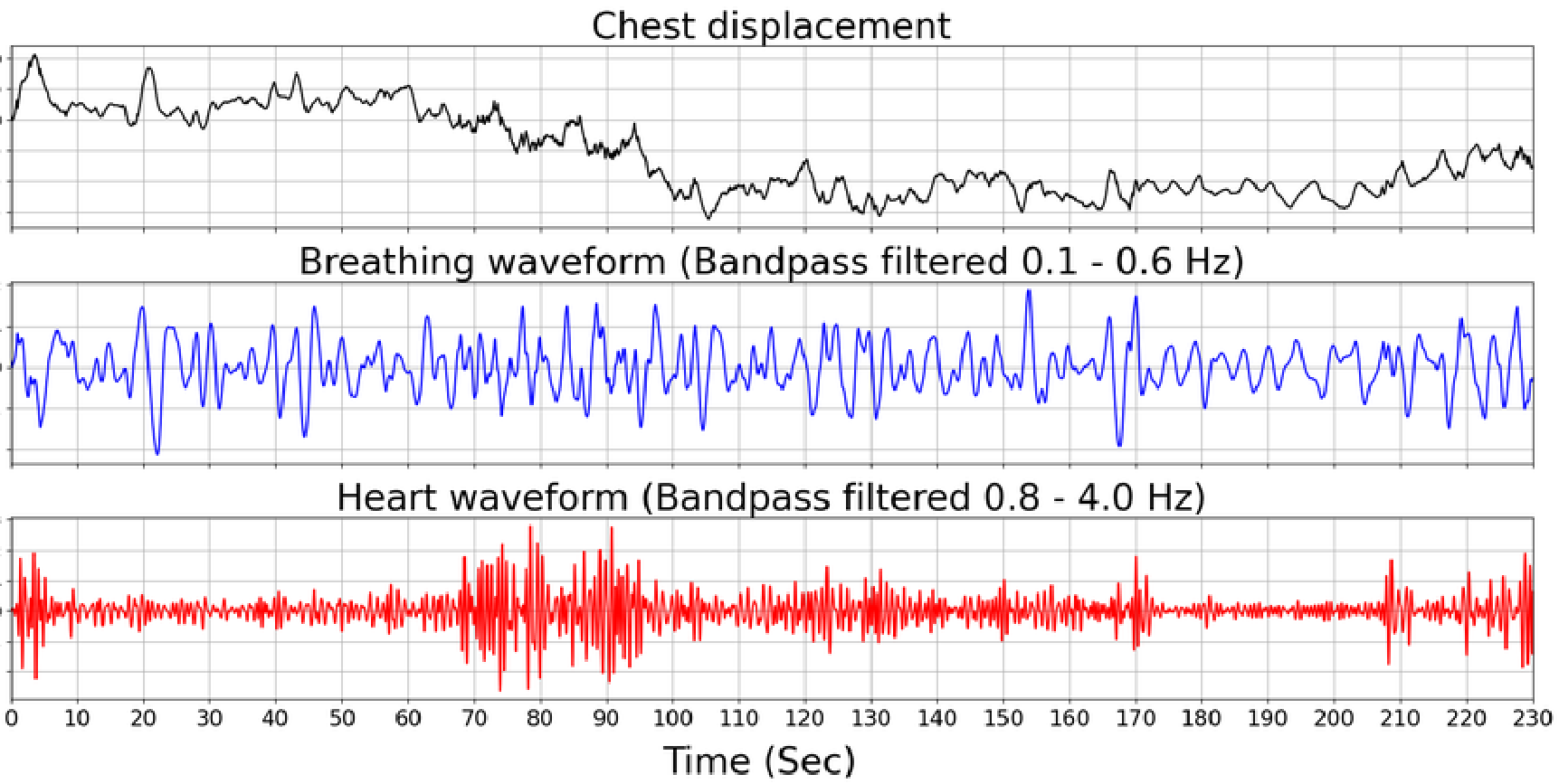}
  \end{center}
  \vspace{-5mm}
  \caption{Three principal waveform are obtained by the mmWave radar in its vital signs configuration: chest displacement (black), breathing waveform (blue) and heartbeat waveform (red).}
  \label{fig:radar_data}
  \vspace{-3mm}
\end{figure}

Texas Instruments IWR1443 is an mmWave radar system that operates within the frequency range of 76-81 GHz, equipped with 4 receivers and 3 transmitters. We adjusted it to capture vital signs configuration within a maximum range of 1.3 meters. This setup allows for the measurement of parameters such as chest displacement of the subject, breathing and heart waveforms and other relevant data not pertinent to this particular study. Employing a Doppler effect, the radar tracks target motion by analyzing the frequency shift in the received waves. The mmWave radar captures chest movements attributed to heart and breathing activity, ranging from 0.5 mm to 4-12 mm, respectively \cite{mikhelson2011remote}. The radar uses a specific vital-sign acquisition firmware which processes a 16-second running window, updating estimates every second, initially calculating distance to the subject for chest displacement assessment. Vital signs extraction involves taking 100 samples of the analog signal within each 50 ms chirp duration. In the firmware, a Fast Fourier Transform (FFT) is performed on these samples to obtain the range profile, from which phase values are computed and analyzed over time. A phase difference operation is conducted on the unwrapped phase by subtracting consecutive phase values, aiding in the enhancement of signal while eliminating any phase drifts \cite {IWR1443}. This configuration generates a data stream at a rate of 20 samples per second, each annotated with its corresponding timestamp in a \textit{csv} file. The three preprocessed streams from the radar used in this study containing the filtered chest displacement of the subject and waveforms for breathing and heartbeat can be shown in Fig. \ref{fig:radar_data}.

\subsection{Setup motivation}

The experiment monitors the subjects engaging in four different activities that encompass varying breathing patterns. The breathing patterns are: normal breathing, reading out loud a text, guided breathing, and breath-holding (apnea). Each breathing pattern spans a duration of 30 seconds and all activities are recorded continuously. These four activities are recorded in three different postures for each user: standing (Pose A), sitting on a chair (Pose B) and lying down on a mat (Pose C).

We selected these four breathing pattern activities because they represent typical scenarios and also simulate emergency situations. The patterns focus on different aspects of the individual breathing, giving specific characteristics such as respiratory rate and motion intensity, which are specially visible in the radar signals.

In normal breathing individuals exhibit innate breathing patterns, that show some regularity but also vary greatly among individuals. Radar signals generated in this state often show consistent and predictable Doppler shifts, corresponding to a semi-regular respiratory motion. In the short span of the breathing pattern collection, the respiratory rate remains relatively stable, following the individual's natural breathing rhythm. The intensity of motion is relatively subtle, reflecting the calm and steady nature of normal breathing, being the normal status in common home scenarios.

When individuals read a text, their breathing pattern becomes irregular, influenced by the content and speech-related movements. This irregularity introduces additional variations in Doppler shifts, creating fluctuations in the radar signals. Head and hand movements, typical while reading, contribute to the intensity of motion during this breathing pattern.

Guided breathing exercises involve following specific rhythmic patterns. This controlled breathing method results in modified and intentional respiratory rhythms. Doppler radar signals exhibit consistent shifts at a fixed frequency during guided breathing, reflecting deliberate variations in breathing rate. These shifts are not only discernible but also visibly apparent. The intensity of motion remains relatively subtle, but more intense than with regular breathing, aligning with the focused and controlled nature of guided breathing exercises.

Apnea or breath holding, represents a distinct pattern with a cessation of breathing, with no discernible Doppler shifts during its occurrence. The lack of motion is most noticeable while seating and lying down, which is particularly pertinent for studying emergency scenarios where an individual does not exhibit breathing movements.

\begin{figure}[h]
  \begin{center}
    \includegraphics*[width=1\columnwidth]{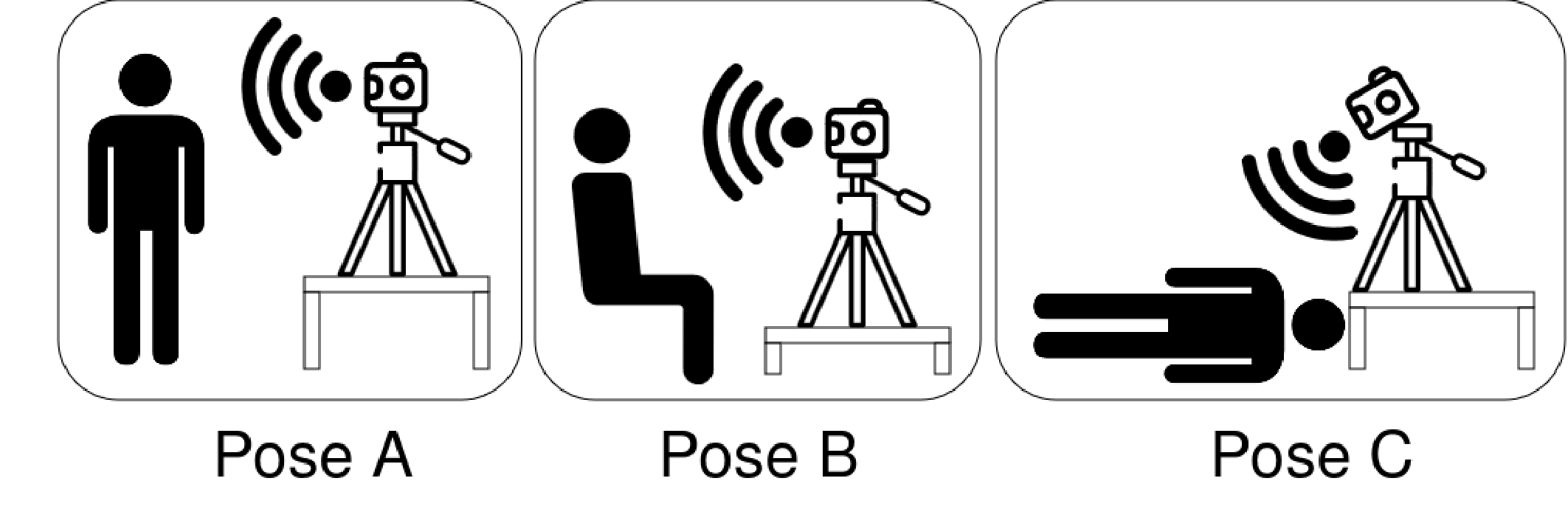}
  \end{center}
    \vspace{-5mm}
  \caption{The three poses: Pose A, standing. Pose B, sitting. Pose C, lying down. Each of the poses stands within a range of 1.3 meters to the mmWave radar and RGB-D Camera.}
  \label{fig:poses}
\end{figure}

The chosen set of three poses (Pose A, Pose B, Pose C) aims to capture the most common stationary positions observed in everyday home scenarios. Each pose exhibits distinctive characteristics of Doppler shift movements, enabling radar analysis to provide unique insights into human movement patterns. A basic scheme of the three poses is presented in Fig. \ref{fig:poses}. First, the standing pose, Pose A, provides a greater range of movement in the upper body, allowing for the detection of a wider variety of Doppler shifts. Due to the inevitable subtle swaying of the standing individuals, it also introduces certain noise affecting at the quality of captured data.

For Pose B, the user is seated in a chair facing the radar. The selection of this pose, especially utilizing the backrest of the chair, was made due to its ability to provide a stable position and a consistent distance for both the radar and camera systems. Furthermore, this posture reflects common household activities such as watching TV, reading, or using a computer, making the study's findings applicable to typical household settings.

Finally, for Pose C, the user lies down on a mat positioned in front of the radar.
The lying-down pose ensures the most stable body position, which yields distinct information during monitoring with the Doppler radar, although it is more challenging in terms of choosing an appropriate positioning of the devices with respect to the user. In this position subjects are generally more relaxed, leading to stable physiological conditions. 

To ensure data consistency, we conducted a thorough review of each recorded pose, promptly re-recording any pose exhibiting artifacts or missing data in breathing activity. This approach guarantees the integrity of the data and ensures that no activities were missing from our dataset.

\subsubsection{Data Recording Protocol}

We collect the data in three different sessions per user, encompassing all four breathing patterns for one specific pose. The recorded time for each pose type is approximately 3.5 minutes, which gives 10.5 minutes for three poses recorded per user. The total duration time of the experiment per user including user measurements, setup and explanation is approximately 25 minutes. 

\begin{figure*}[ht!]
  \begin{center}
    \includegraphics*[width=0.95\textwidth]{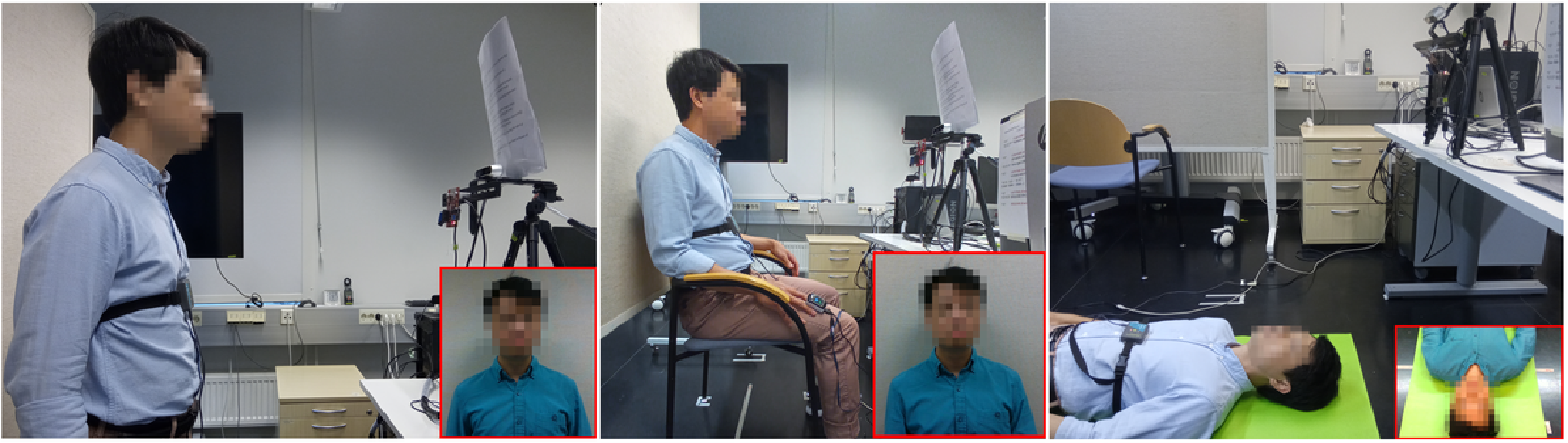}
  \end{center}
  \caption{Experimental setup for bio-movement radar analysis: Pose A (left) shows the subject standing 70 cm from the radar for breathing tasks. In Pose B (center), the subject is seated, with torso movement restricted by the chair, altering Doppler shifts. Pose C (right) has the subject lying down, executing the same tasks without torso constraints. The red square depicts the real image captured by the camera.}
  \label{fig:poses_rgb}
  \vspace{-3mm}
\end{figure*}

The preparation of the three poses aims to reduce individual differences among users and prevent potential biases related to height estimation based on camera angles. For this we individually adapt the heights of both the camera and radar system for each user. Pose A records the user performing the four breathing tasks standing up in front of the tripod. The preparation for this pose involves the manual height adjustment of our tripod to a fixed position as well as our adjustable desk. This positioning guarantees that the RGB image frame captured the user's head slightly above the frame, as illustrated in Fig. \ref{fig:rgb_d}. To reduce the dynamic range of the cameras, a non-reflective panel was strategically placed behind the user. This approach also aids in reducing reflections, ensuring clear imaging, and minimizing the dynamic range of both RGB and depth cameras.
The user was instructed to stand at a fixed distance of 70 cm away from the radar, which was facilitated by floor stickers indicating the correct placement.
In Fig. \ref{fig:poses_rgb} (left), we provide a visual representation of the correct subject placement and the resulting image captured by the RGB stream.

Upon completion of the first pose, the recording was halted, and we inspected the recorded radar information ensuring that there were no instances of corrupted data.
Pose B records the user sitting on a chair in front of the tripod.
To maintain consistency and ensure unbiased data collection, similar to the previous setup, we adjusted the tripod and table to a new fixed height. 
In Pose B, the user assumes a seated position, facing the radar, as depicted in Fig. \ref{fig:poses_rgb} (center). The precise placement of the chair was indicated on the ground by stickers, ensuring uniform positioning across participants.
In this seated posture, users were instructed to sit back in a relaxed manner, utilizing the chair's backrest. This specific pose was chosen strategically to restrict chest movement, enabling clearer detection of respiratory patterns by the radar. Upon the completion of all three poses, a secondary monitoring session for blood pressure was conducted. 

Pose C records the user lying down in front of the tripod. Subsequently a non-reflective mat was used. The camera angle was carefully adjusted to encompass the user's face and chest entirely, as demonstrated in Fig. \ref{fig:poses_rgb} (right). 
This adjustment was undertaken to capture a more comprehensive range of visual information during the data collection process. It is noteworthy that the resulting image for Pose C is presented as a 180º rotated video.

\begin{figure}[h]
  \begin{center}
    \includegraphics*[width=1\columnwidth]{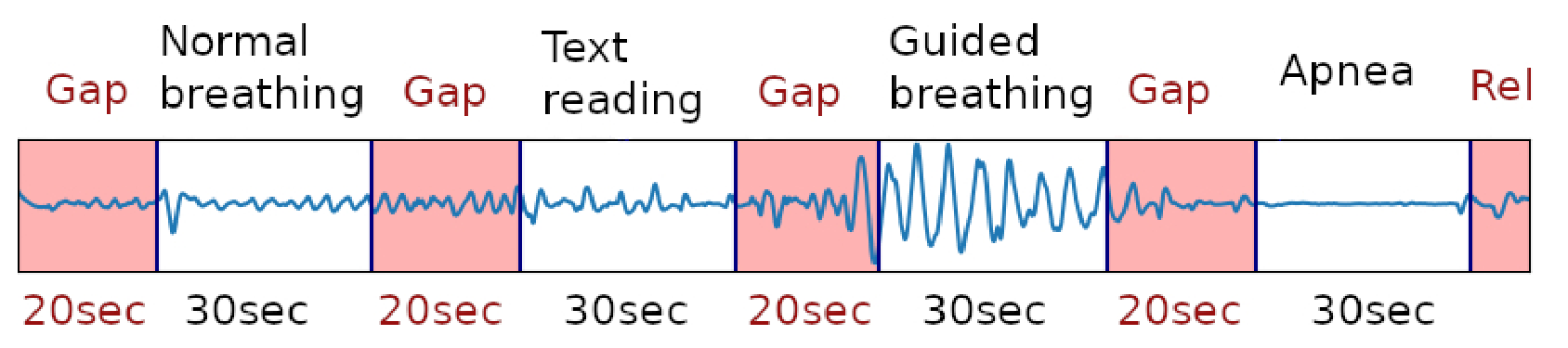}
  \end{center}
    \vspace{-3mm}
  \caption{Data protocol with the four consecutive activities per pose. Gap time segments serve as intermediate time sections to prepare the user for the next task. In this example, the blue line represents the breathing pattern measured by the radar.}
  \label{fig:protocol}
  \vspace{-3mm}
\end{figure}

The four breathing activities are recorded uninterruptedly for every pose. Each breathing pattern lasts 30 seconds separated with a gap time of 20 seconds. The 20 second intervals serve a dual purpose: enhancing task separation during post-processing and providing a time frame for verbal advises from the staff, reminding the user the guidelines for the forthcoming task. The duration of the exercises is regulated by a global timer, which emits an audible alert to notify the user when a task commences or concludes. When the recording starts, the timer emits a beep indicating the starting of a first gap of 20 seconds in which we reminded the user the rules for normal breathing task. A second beep by the timer indicates the starting of the normal breathing. A visualization of the data collection protocol time line for each session (encompassing one pose) can be seen in Fig. \ref{fig:protocol}. The figure depicts a respiration rate measured by the radar in blue as a guide for better understanding. 

The timer registers the initiation and duration of each breathing pattern, with updates occurring on a per-second basis. Throughout the 20 seconds gap intervals, it categorizes the recorded activity as idle. These timestamped entries are compiled into a separate file, intended to aid in the segmentation and differentiation of various activities during data preprocessing.

The first breathing exercise consists in breathing normally in front of the radar upon receiving the aural alert. The participant initiates the task by standing up and maintaining an upright standing position at a fixed distance of the radar, keeping their feet shoulder-width apart and their arms relaxed by their sides. Throughout the 30 second duration, the participant breathes naturally and avoids any intentional alterations to their breathing rhythm or depth, focusing on maintaining a steady posture and minimizing unnecessary body movements.

The second breathing pattern involves the participant reading a predetermined text consisting of 100 words aloud. The participant is given the freedom to read at their own pace or speed. However, there are specific conditions related to task completion. If the participant has not finished reading the entire text before the aural alert signals the end of the 30-second duration, the staff will instruct them to stop reading at that point. Conversely, if the participant reads at an exceptionally fast pace, completing the entire text within the 30-second time frame, they are instructed to begin reading from the beginning once again.
These conditions ensure consistency and allow for appropriate assessment of the reading task within the given time constraint.

The third breathing activity involves a guided breathing exercise. Participants are instructed that they will hear a specific aural pattern, different from the starting or ending task, and are required to synchronize their breathing with it. 
The pattern consists of two low-pitched sounds at 440 Hz, followed by two high-pitched sounds at 880 Hz. Each sound has a duration of 1 second, resulting in a total pattern duration of 4 seconds. The task is designed to last 30 seconds, and it begins with the low-pitched sounds. Therefore, participants are instructed to exhale during the two low-pitched sounds and inhale during the two high-pitched sounds. Given the total duration of the task, participants will perform eight exhalations and seven inhalations in total,  This guided breathing task provides a structured approach to regulate the participants breathing patterns and facilitates data collection and analysis.

The fourth breathing pattern in the protocol is holding the breath for a duration of 30 seconds, while also maintaining a steady posture. This task holds particular significance in Pose C, where participants are lying down on a mat and mimics the situation of a lying body that has stopped breathing. Participants are informed that if they find it challenging to hold their breath for the entire 30 seconds duration, they should release their breath and promptly resume holding the breath again to maximize the time of chest immobility within the designated task duration.
By emphasizing the importance of maintaining breath-holding and stability, this task aims to capture different physiological responses and detect any variations in radar readings.

\begin{figure}[h]
  \begin{center}
    \includegraphics*[width=1\columnwidth]{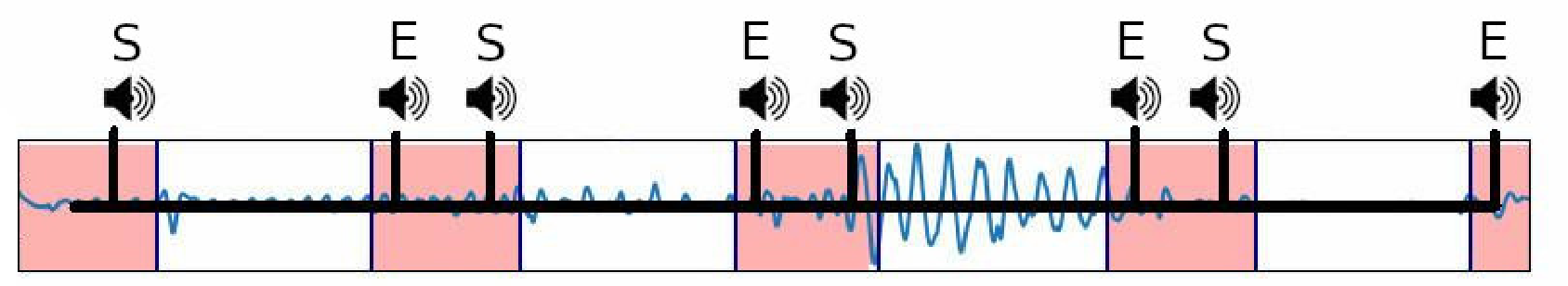}
  \end{center}
  \vspace{-3mm}
  \caption{Aural alert pattern. Each vertical black line indicates an aural alert. The aural alerts inform the user of the starting (S) and ending (E) of an exercise. The alerts are specifically located outside the exercise measurement zones to ensure that the time zones only measure the specific task, thus, avoiding actions such as apnea inhalation, apnea exhalation and silence before reading.}
  \label{fig:aural_alert}
  \vspace{-3mm}
\end{figure}

It was observed that participants often exhibited varying delay times when initiating breathing patterns, such as starting to read a text. While some individuals immediately commenced the task upon the timer's activation, others displayed delays of one or even two seconds before emitting any sound.
To mitigate this issue, a system was implemented to ensure that exercises were initiated precisely when the exercise timer commenced. Participants were instructed to start each task upon hearing a distinct starting aural alert and to stop the task upon hearing the corresponding ending alert. The starting alert was designed to sound two seconds before the timer task initiated, while the ending alert occurred two seconds after the timer marked the task's conclusion. Consequently, every task performed by the user was initiated and concluded within the gap time, guaranteeing that the total exercise recorded time accurately reflected the duration of the respective task. The aural alerts served as a sanity check and were not included in the timer recording file, thus ensuring that only the timer timestamps were utilized for subsequent data analysis. A schematic representation of the starting and ending sound patterns for the users can be found in Fig. \ref{fig:aural_alert}

\subsection{Hardware setup}

The daily setup during the data collection includes the synchronization of the two PCs involved in the data collection, annotation of temperature, humidity and lumens of the room and a record of 30 seconds of radar of the empty laboratory for comparison purpose.
The data analysis was done using a computer that integrates an AMD Ryzen 7 5800 8-Core processor and an NVIDIA GeForce RTX 3060 running on Linux. We used Python 3.9 as the programming language.


\section{Data processing methodology}

For every user, three sessions were recorded. Each session registers data from camera and radar of one pose with the four breathing patterns sequence activities.
The raw data per session includes two video streams, one containing the RGB format and another containing the depth information and three preprocessed waveforms from the radar outputing chest displacement of the subject and estimation of heartbeat and breathing signals.

A supplementary \textit{csv} file containing timestamps for each instance of respiratory activity within a video pose is included. It aids in the separation of activities within the signals and ensures synchronization of data collected from different PCs.

\subsection{Dataset preprocessing}

\subsubsection{Signal extraction}
For both camera and radar data, all preprocessing is conducted independently of the model development and stored in a separate dataframe. The preprocessing of the RGB stream involves the extraction of video-based biosignals, in our particular case, remote photoplethysmography signals (rPPG). Analogous to photoplethysmography signals obtained with pulse oximeters, rPPGs, are biosignals obtained in a non-invasive, contactless method that uses a video camera and ambient light to detect variations in skin color due to blood flow, reflecting the complex interactions of light with skin properties\cite{huang2023challenges}.

From the RGB stream video input, we derive three different (rPPG) signals using the Face2PPG framework \cite{casado2023face2ppg}, an unsupervised rPPG acquisition method that involves applying a facial detection and segmentation algorithm to each frame of the video stream. Subsequently Face2PPG employs signal processing and a chrominance-based method (CHROM) \cite{de2013robust} to derive the rPPG signals for the entire face, as well as for the left and right regions separately, denoted as rPPG, rPPG-Left and rPPG-Right, respectively. CHROM method applies simple linear combinations of RGB channels and projects them onto two orthogonal chrominance vectors X and Y \cite{benezeth2018remote}. 

\begin{figure}[h]
  \begin{center}
    \includegraphics*[width=1\columnwidth]{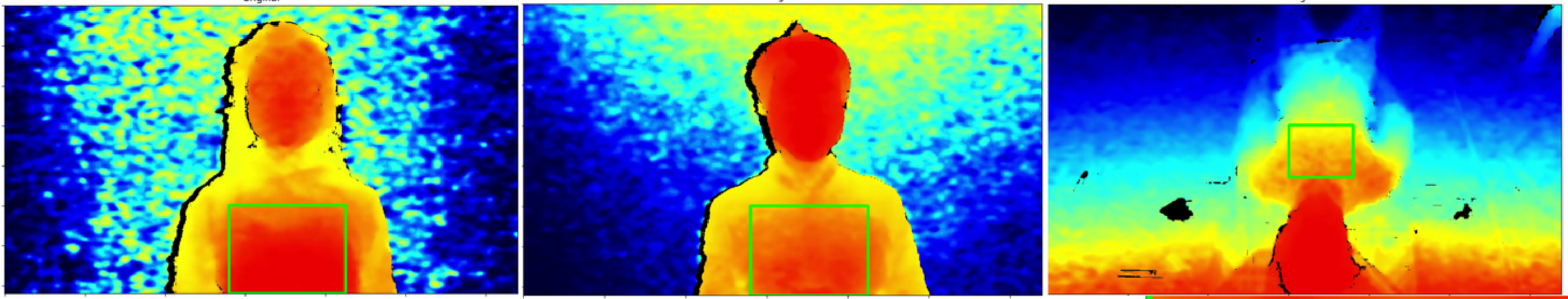}
  \end{center}
  \vspace{-3mm}
  \caption{Central chest patch (green square) manually placed from the depth images where the scalar distance is calculated.}
  \label{fig:Depth patches}
\end{figure}

In addition to the RGB stream, the video depth stream is used to derive a distance signal. For each pose, we designate a central region of the video focused on the chest area aiming to maximize the area where respiration can be observed. 
Since all individuals are recorded at a constant distance and under identical conditions, the chest position remains consistent throughout the video and across subjects, as depicted in the example in Fig. \ref{fig:Depth patches}. The RGB information of this patch is then extracted for every frame of the video and converted into 8-bit pixel values. The values for each pixel in the patch are then averaged, resulting a scalar value and stored as a separate signal.

In its vital signs configuration mode, the mmWave radar derives several signals, from which only three waveforms are utilized. Chest displacement is determined by monitoring changes in the phase of the Frequency-Modulated Continuous Wave (FMCW) signal over time at the target range bin. Subsequently, these waveforms are stored in the preprocessed dataframe for subsequent feature extraction. Since for adults, typical vital sign parameters for breathing rate amplitude ranges from 1 to 12 mm with a frequency between 0.1 and 0.5 Hz, and the heart rate amplitude ranges from 0.1 to 0.5 mm with a frequency between 0.8 and 2 Hz. Subsequently, to extract the heartbeat waveform (HW) and breathing waveform (BW), two bandpass filters are utilized: one ranging from 0.1 to 0.6 Hz for breathing and another from 0.8 to 4.0 Hz for the heartbeat.

\begin{figure*}[h]
  \begin{center}
    \includegraphics*[width=0.88\textwidth]{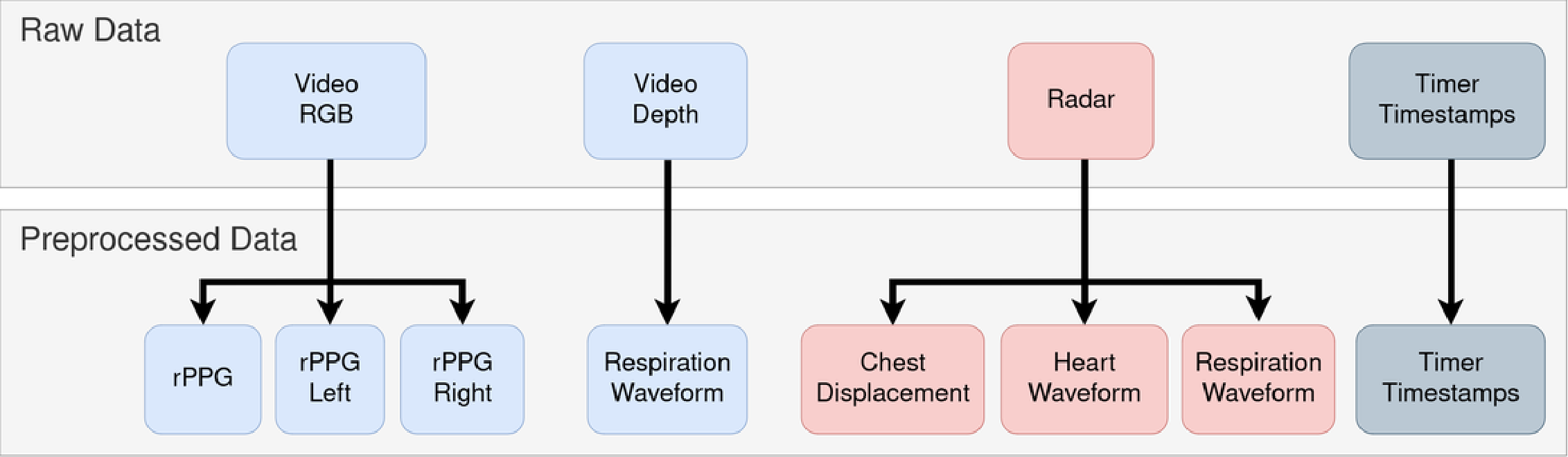}
  \end{center}
  \caption{List of seven preprocessed signals used for our study. We include the timestamps from the timer to help in activity separation.}
  \label{fig:raw_signals}
\end{figure*}

The seven signals analyzed in our study are illustrated in Fig. \ref{fig:raw_signals}. These include three rPPG streams extracted from the RGB camera, a respiration waveform signal derived from the depth camera, as well as chest displacement, heart and breath waveforms obtained from radar data. Additionally, a timestamp is included to aid in synchronization and activity separation. Since both radar and camera operate at fundamentally different sample rates, all signals resampling to the a lowest 20 Hz frequency using linear interpolation.

\subsubsection{Activity separation and window segmentation}
Each preprocessed signal file comprises a continuous recording of the four breathing pattern activities of 30 seconds followed by 20 seconds idle time. Activity separation is achieved by referencing the timestamps in the timer file and eliminating all 20 seconds gap time. This results in a total of 600 recorded activities (50 users $\times$ 3 poses $\times$ 4 activities). Subsequently, each preprocessed 30-second breathing pattern is segmented into discrete ten-second windows following common approaches in the literature \cite{putra2017window}. 
The windows incorporate a one-second step shift between consecutive windows meaning that two consecutive windows overlap in all signals during 9 seconds. The windowing process for the whole dataset results in over 12,000 windows. After the segmentation, we labels every single window with user, pose and activity. 

\subsection{Feature Extraction}
For each window we derive an extensive set of features, including statistical features, fractal analysis features, and complex analysis features. From signals related to breath, and heart, we compute specific breath features based on their amplitude and variability using SciPy library \cite{scipy} and HeartPy libraries \cite{van2019heartpy}. 

\subsubsection{Statistical, fractal and entropy features}
In a similar manner related literature, we extract from each signal and window seven well-known features \cite{makela2021introducing} (\textit{mean}, \textit{median}, \textit{SD}, \textit{max}, \textit{min}, \textit{25th percentile,} and \textit{75th percentile}), and five additional statistical features \cite{casado2023depression} (\textit{median absolute deviation}, \textit{four interquartiles}).  The fractal properties of time series data involves assessing patterns with self-similarity across different scales. For this purpose, we calculate \textit{Katz} and \textit{Higuchi} fractal dimensions \cite{raghavendra2009note} to quantify irregularities in a curve over a time series, along with \textit{detrended fluctuation analysis} (DFA) to explore the presence of long-range correlations in the data. \textit{Hjorth} parameters, including mobility and spectral entropy, are also calculated.

\subsubsection{Physiological Features}
In addition to those features, we add two subsets of physiological from heart and breath waveforms. Heart and heart rate variability (HRV) analyses were calculated using the HeartPy libraries on signals from cardiac frequencies. The extracted features encompass Heart Rate (HR), Beats per Minute (BPM), Inter-beat Interval, and various calculations based on normal-to-normal (NN) intervals. These NN interval computations include SD and the percentage of adjacent NN intervals differing by more than 20 ms and 50 ms. 
Additionally we calculate interbeat interval (IBI), differences between beats intervals (pNN20, pNN50), Poincare analysis, frequency domain components (VLF, LF, HF, LF/HF ratio). Finally, employing the HRV Analysis library \cite{caridade2019development}, we conducted a comprehensive examination of short-term heart rate variability, encompassing the coefficient of variation, as well as key metrics derived from NN intervals, namely \textit{mean}, \textit{max}, \textit{min}, and \textit{SD} of heart rates.

An analysis of breath-related features was conducted, focusing on derived parameters from respiratory data. The calculation of breathing parameters involved consideration of various phases within the respiratory cycle, including SD, mean, maximum, and minimum amplitude between intervals of exhalation and inspiration. Additionally, the mean and SD of amplitudes in the respiratory signal were computed, providing insights into variability. Furthermore, the respiratory rate was determined, offering a quantification of the number of breaths per minute. Notably, key features such as the mean duration of NN intervals during both the expiratory and inspiratory phases of the respiratory cycle were also assessed. 

\section {Regression and classification models}
The extracted features are used to feed a selection of machine learning algorithms.
We have opted for a set of four regressors and four classifiers, which have been implemented in Python using the scikit-learn library. For the regressors, we have chosen XGBoost (\textit{XGB}), Random Forest (\textit{RF}), Support Vector Machine Regressor (\textit{SVR}), and Adaboost (\textit{AB}). As for the classifiers, our selection includes XGBoost (\textit{XGB}), Random Forest (\textit{RF}), Adaboost (\textit{AB}), and Multilayer Perceptron (\textit{MLP}). For the hyperparameters, XGB is used with 1500 estimators. SVR is left with a coeficient 1.0 and epsilon 0.2. AB is used with 100 estimators. MLP uses a solver adam with 3000 max iteration and a initial learning rate of 0.04. RF is used with the default configuration.

\section {Evaluation}
We define several testing and validation protocols for a set of classification (pose, breathing pattern) and regression (age, weight, height) tasks. In our evaluations, the poses, breathing patterns and the combinations of both pose and activities (\textit{posActs}) are defined as follows: 

\begin{itemize}
   \item \textbf{Poses:} \textit{Pose A, standing (PA), Pose B, sitting (PB) and Pose C, laying down (PC). }

   \item \textbf{Breathing patterns:} \textit{Normal breathing (A1), Reading (A2), Guided breathing (A3) and Apnea (A4).}

   \item \textbf{Combinations (posActs):} \textit{Standing and Normal Breathing (C1), Standing and Reading (C2), Standing and Guided Breathing (C3), Standing and Apnea (C4), Sitting and Normal Breathing (C5), Sitting and Reading (C6), Sitting and Guided Breathing (C7), Sitting and Apnea (C8), Lying Down and Normal Breathing (C9), Lying Down and Reading (C10), Lying Down and Guided Breathing (C11), Lying Down and Apnea (C12).}
 \end{itemize}

\subsection{Evaluation Protocols}
We propose three different protocols for partitioning the dataset for evaluation, employing the same evaluation metrics across all of them. Each protocol encompasses regression analysis for biometric characteristics and classification for differentiating between pose, breathing patterns and \textit{posActs}.

\subsubsection{Leave-one-subject-out LOSO cross-validation} 
This involves training models utilizing a LOSO cross-validation approach, where each training iteration utilizes data from 49 users and evaluates the models with the unseen data of the remaining user presenting the results as the mean of the predictions for every individual window. As a general metric we summarized the results in the mean of 50 training models. The results of this first protocol are presented in four different stages.

\subsubsection{K-fold cross-validation} 
The dataset is partitioned into five folds, each comprising ten users ensuring sex  balance across all five folds. The division by user is outlined below:

\begin{itemize}
    \item \textbf{Fold I}: \textit{01, 02, 03, 04, 05, 06, 07, 08, 09, 10}
    \item \textbf{Fold II}: \textit{11, 12, 13, 14, 15, 16, 17, 18, 19, 20}
    \item \textbf{Fold III}: \textit{21, 22, 23, 24, 25, 26, 27, 28, 30, 31}
    \item \textbf{Fold IV}: \textit{29, 32, 33, 34, 35, 36, 38, 39, 41, 42}
    \item \textbf{Fold V}: \textit{37, 40, 43, 44, 45, 46, 47, 48, 49, 50}
\end{itemize}

\subsubsection{Train-validation-test data split}
This protocol splits the data in three subsets: training, validation and testing sets. The training set contains 30 users (folds I, II and III), validation (fold number IV) and test (fold number V) ten users each. 

\subsection{Evaluation Metrics}
The performance of the regressors (tasks for age, weight and height) is evaluated using using Mean Absolute Error (MAE), Root Mean Square Error (RMSE), and Mean Absolute Percentage Error (MAPE).  For classification, we measure the effectiveness with four metrics (accuracy, precision, recall and F1 score) and for selected cases, also a confusion matrix. 

\section{Experimental Results}
We provide baselines results for each sensor modality separatedly (RGB-D Camera and mmWave Radar). This is achieved using a simple protocol comprising eight statistical features, a model using extended statistical features and a model trained additionnally with physiological features. Moreover, we provide a stratification experiment training over specific poses and breathing patterns. Finally we compute results for joint modalities, including two additional validation strategies (k-fold and data-split). 

\subsection{Simple protocol baseline}
We establish a straightforward baseline protocol on just the radar modality, employing a simplified approach for both classification and regression tasks, aimed at gaining initial insights. Our initial baseline protocol incorporates just eight statistical features extracted from the three radar signals: \textit{max}, \textit{min}, \textit{mean}, \textit{SD}, \textit{kurtosis}, \textit{skewness}, \textit{variance} and \textit{median} per window and signal, resulting in a total of 24 features. Using these basic features, we utilize a Random Forest classification, focusing on pose, activity, sex and a 12-class \textit{posAct}. Additionally, we present the results of MAE with a Random Forest regressor for age, weight, and height. The results, presented in Table \ref{tab:rad_basic}, are divided into two sections; classification with accuracy on the left and regression task with MAE error on the right.

\begin{table}[ht!]
\renewcommand{\arraystretch}{1.1}
\setlength{\tabcolsep}{0.7em}

\caption{First baseline: Classifier and Regressor (24 radar-based features) using Random Forest for all 50 participants}
\label{tab:rad_basic}
\centering
\begin{tabular}{r|cccc|ccc}

\multicolumn{1}{c|}{}& \multicolumn{4}{c|}{Classifiers (Accuracy)}& \multicolumn{3}{c}{Regressors (MAE)} \\

User &Pose  &Activity &Sex  &posAct &Age  &Height &Weight         \\
\hline
1   &0.71  &0.51   &0.35  &0.36   &8.15 	 &12.39    &27.36     \\
2   &0.68  &0.77   &0.55  &0.46   &5.23 	 &5.53 	   &10.01     \\
3   &0.74  &0.71   &0.44  &0.55   &7.17 	 &4.67 	   &6.36      \\
4   &0.64  &0.66   &0.76  &0.39   &1.79 	 &9.98 	   &16.38     \\
5   &0.77  &0.55   &0.52  &0.38   &5.19 	 &7.70 	   &14.10     \\
6   &0.70  &0.53   &0.59  &0.37   &8.78 	 &8.27 	   &5.97      \\
7   &0.77  &0.72   &0.72  &0.64   &2.72 	 &5.31 	   &7.14      \\
8   &0.35  &0.60   &0.74  &0.33   &5.74 	 &1.46 	   &8.39      \\
9   &0.77  &0.47   &0.50  &0.41   &36.85     &21.05    &24.84     \\
10  &0.74  & 0.69  & 0.25 & 0.55  &3.33 	 &4.39 	   &7.51      \\
11  &0.59  & 0.78  & 0.70 & 0.52  &1.19 	 &3.03 	   &10.14     \\
12  &0.62  & 0.60  & 0.25 & 0.38  &3.71 	 &4.08 	   &6.17      \\
13  &0.61  & 0.55  & 0.89 & 0.40  &5.82 	 &21.00    &11.71     \\
14  &0.67  & 0.59  & 0.12 & 0.44  &3.98 	 &12.34    &9.15      \\
15  &0.71  & 0.66  & 0.34 & 0.53  &4.07 	 &2.84 	   &5.44      \\
16  &0.75  & 0.62  & 0.26 & 0.43  &4.51 	 &11.91    &16.94     \\
17  &0.71  & 0.58  & 0.87 & 0.46  &1.85 	 &4.82 	   &23.26     \\
18  &0.70  & 0.59  & 0.79 & 0.43  &6.05 	 &6.04 	   &11.89     \\
19  &0.62  & 0.48  & 0.53 & 0.29  &11.99     &6.67 	   &42.39     \\
20  &0.86  & 0.83  & 0.74 & 0.71  &3.49 	 &16.87    &8.07      \\
21  &0.81  & 0.50  & 0.72 & 0.58  &2.41 	 &3.25 	   &7.93      \\
22  &0.69  & 0.69  & 0.00 & 0.53  &1.84 	 &17.07    &8.70      \\
23  &0.84  & 0.59  & 0.25 & 0.49  &2.73 	 &11.22    &15.22     \\
24  &0.73  & 0.71  & 0.76 & 0.57  &2.34 	 &11.84    &11.63     \\
25  &0.82  & 0.55  & 0.52 & 0.45  &1.26 	 &6.15 	   &5.60      \\
26  &0.75  &0.38   &0.59 &0.32    &1.99 	 &4.61 	   &7.17      \\ 
27  &0.67  &0.51   &0.85 &0.32    &2.95 	 &9.69 	   &14.76     \\ 
28  &0.70  &0.72   &0.72 &0.57    &3.21 	 &4.98 	   &7.23      \\ 
29  &0.86  &0.56   &0.98 &0.59    &1.91 	 &13.19    &9.53      \\ 
30  &0.67  &0.67   &0.21 &0.45    &3.70 	 &12.33    &18.20     \\
31  &0.86  &0.52   &0.86 &0.47    &7.50 	 &4.72 	   &23.94     \\ 
32  &0.71  &0.84   &0.57 &0.67    &2.66 	 &7.94 	   &17.44     \\ 
33  &0.60  &0.60   &0.44 &0.36    &5.62 	 &7.66 	   &10.38     \\ 
34  &0.47  &0.65   &0.37 &0.36    &11.11     &4.99 	   &7.33      \\ 
35  &0.58  &0.42   &0.60 &0.25    &7.38 	 &2.22 	   &19.30     \\ 
36  &0.75  &0.50   &0.90 &0.30    &11.26     &3.12 	   &8.49      \\ 
37  &0.76  &0.55   &0.68 &0.49    &4.70 	 &4.97 	   &5.72      \\ 
38  &0.72  &0.81   &0.79 &0.66    &3.41 	 &15.67    &9.76      \\ 
39  &0.81  &0.68   &0.33 &0.52    &1.27 	 &8.08 	   &4.50      \\ 
40  &0.73  &0.69   &0.60 &0.58    &5.96 	 &14.65    &14.13     \\
41  &0.66  &0.70   &0.66 &0.46    &6.43 	 &5.90 	   &11.13     \\ 
42  &0.77  &0.59   &0.58 &0.41    &1.62 	 &16.39    &23.48     \\ 
43  &0.62  &0.64   &0.22 &0.47    &6.37 	 &7.38 	   &7.29      \\ 
44  &0.72  &0.74   &0.25 &0.57    &8.43 	 &6.35 	   &8.06      \\ 
45  &0.74  &0.76   &0.50 &0.64    &14.04     &12.48    &14.39     \\
46  &0.78  &0.73   &0.39 &0.56    &6.84 	 &8.45 	   &6.26      \\ 
47  &0.68  &0.62   &0.31 &0.46    &7.77 	 &8.30 	   &9.80      \\ 
48  &0.62  &0.59   &0.57 &0.42    &14.48     &17.98    &16.71     \\
49  &0.78  &0.55   &0.46 &0.46    &1.53 	 &2.96 	   &7.31      \\ 
50  &0.62  &0.69   &0.48 &0.46    &4.00 	 &6.79 	   &9.64      \\ 
\hline 
Mean  &\textbf{0.70}  &\textbf{0.62}  &\textbf{0.54}  &\textbf{0.47}  &\textbf{5.77}  &\textbf{8.63}  &\textbf{12.28}  \\
SD   &0.09           &0.10           &0.23           &0.11           &5.51           &4.97           &7.12            \\
\hline
\multicolumn{1}{c|}{}& \multicolumn{4}{c|}{Random guess}& \multicolumn{3}{c}{Naive Regressor MAE}                      \\
 &\textbf{0.33} &\textbf{0.25} &\textbf{0.50} &\textbf{0.08} &\textbf{4.65} &\textbf{7.73} &\textbf{11.20}             \\
\end{tabular}
\end{table}

Classification models for pose identification performs better than random guess. Specifically, the classifier for pose achieves a mean accuracy of 70\% compared to 33\% expected with random guessing. The main finding is that the relative low SD shows consistent performance across subjects, suggesting that basic statistical information from the radar sufficiently generalized for unseen users, given that all users' data was collected in similar conditions.

Activity classification achieves similar result; 62\% compared with a 25\% of the random guess. However, the SD indicates variability in performance across subjects, with some subjects potentially performing better or worse than others. Sex classification performs worse but still better than random guess with a mean accuracy of 54\% compared of 50\% random guess of a two class classifier. The results present high SD, highlighting substantial variability in performance across subjects. Finally, the 12-class \textit{posAct} mean accuracy of 47\% is significantly higher than the random guess of 8\%, with relatively consistent results among users, indicating that the model performs better than random chance on average for this task.

The regression analysis conducted on biometric and physiological data reveals suboptimal performance, where e.g. the age prediction task shows worse performance (MAE 5.77) than a naive regressor that guesses the average age (MAE 4.65), suggesting that radar features does not provide enough information to correctly reflect physiological characteristics. The considerable variability in errors across subjects further indicates that age estimation based solely on these simple statistical features is unreliable. The regression model for height and age shows similar results.

\subsection{Extended statistical features}
This experiment includes classification tasks in separate modalities utilizing statistical features from both radar and camera devices, corresponding to 118 radar features, and 156 camera features. The results are computed for four different classifiers, and the findings are presented in Table \ref{tab:clas_stat}. 

\begin{table}[ht!]
\renewcommand{\arraystretch}{1.1}
\setlength{\tabcolsep}{0.25em}
\caption{Classification with extended statistical features}
\label{tab:clas_stat}
\centering
\scalebox{0.85}{
\begin{tabular}{l|ccccc|cccc}
&\multicolumn{5}{c}{Radar features (118)} & \multicolumn{4}{c}{Camera features (156)} \\
Task      & Classifier  & Accuracy  & Precision & Recall & F1& Accuracy  & Precision & Recall & F1\\
\hline
\multirow{4}{*}{Pose}
&  XGB 	 & \textbf{0.77}  &0.77  &0.79 &0.76   &\textbf{0.76} &0.76 &0.78 &0.74  \\
&  RF    &0.74  &0.74 	&0.76 &0.74 &0.74 &0.74 &0.77 &0.73 \\
&  AB    &0.71  &0.71 	&0.72 &0.70 &0.69 &0.69 &0.71 &0.66 \\
&  MLP	 &0.62  &0.62 	&0.63 &0.59  &0.62 &0.62 &0.58 &0.55 \\
\hline                                                                                                                       
\multirow{4}{*}{Activity} 
& XGB	  & \textbf{0.71}  &0.71  &0.74 &0.70 &\textbf{0.69} &0.69 &0.73 &0.68 \\
& RF	  &0.70  &0.70 	&0.73 &0.69 &0.65 &0.65 &0.69 &0.63  \\
& AB	  &0.64  &0.64 	&0.67 &0.63 &0.65 &0.65 &0.68 &0.64  \\
& MLP	  &0.49  &0.49 	&0.51 &0.45 &0.27 &0.25 &0.08 &0.11 \\
\hline                                                                                                                       
\multirow{4}{*}{Sex}     
& XGB    &0.53 &0.27 &0.50 &0.34 &0.52 &0.26 &0.50 &0.33  \\
& RF	 &0.54 &0.27 &0.50 &0.33 &\textbf{0.54} &0.27 &0.50 &0.34 \\
& AB	 & \textbf{0.55} &0.28 &0.51 &0.35&0.52 &0.26 &0.50 &0.33 \\
& MLP	 &0.30 &0.20 &0.36 &0.22 &0.25 &0.15 &0.30 &0.17 \\
\hline                                                                                                                     
\multirow{4}{*}{posAct}  
& XGB     & \textbf{0.57} &0.58 &0.59 &0.54 &\textbf{0.51} &0.51 &0.52 &0.47  \\
& RF	  & \textbf{0.57} &0.57 &0.58 &0.53 &0.48 &0.48 &0.47 &0.43  \\
& AB	  &0.30 &0.30 &0.26 &0.23 &0.27 &0.27 &0.22 &0.20  \\
& MLP	  &0.21 &0.21 &0.12 &0.12 &0.09 &0.09 &0.01 &0.02  \\                               
\end{tabular}}                                                                    
\end{table}

The results indicate an improvement compared to the initial baseline for all classifiers. The best classifier seems to be XGB although the performance compared with the alternatives is very similar and not consistent across tasks.  

\begin{figure}[h!]
  \begin{center}
    \includegraphics*[width=1\columnwidth]{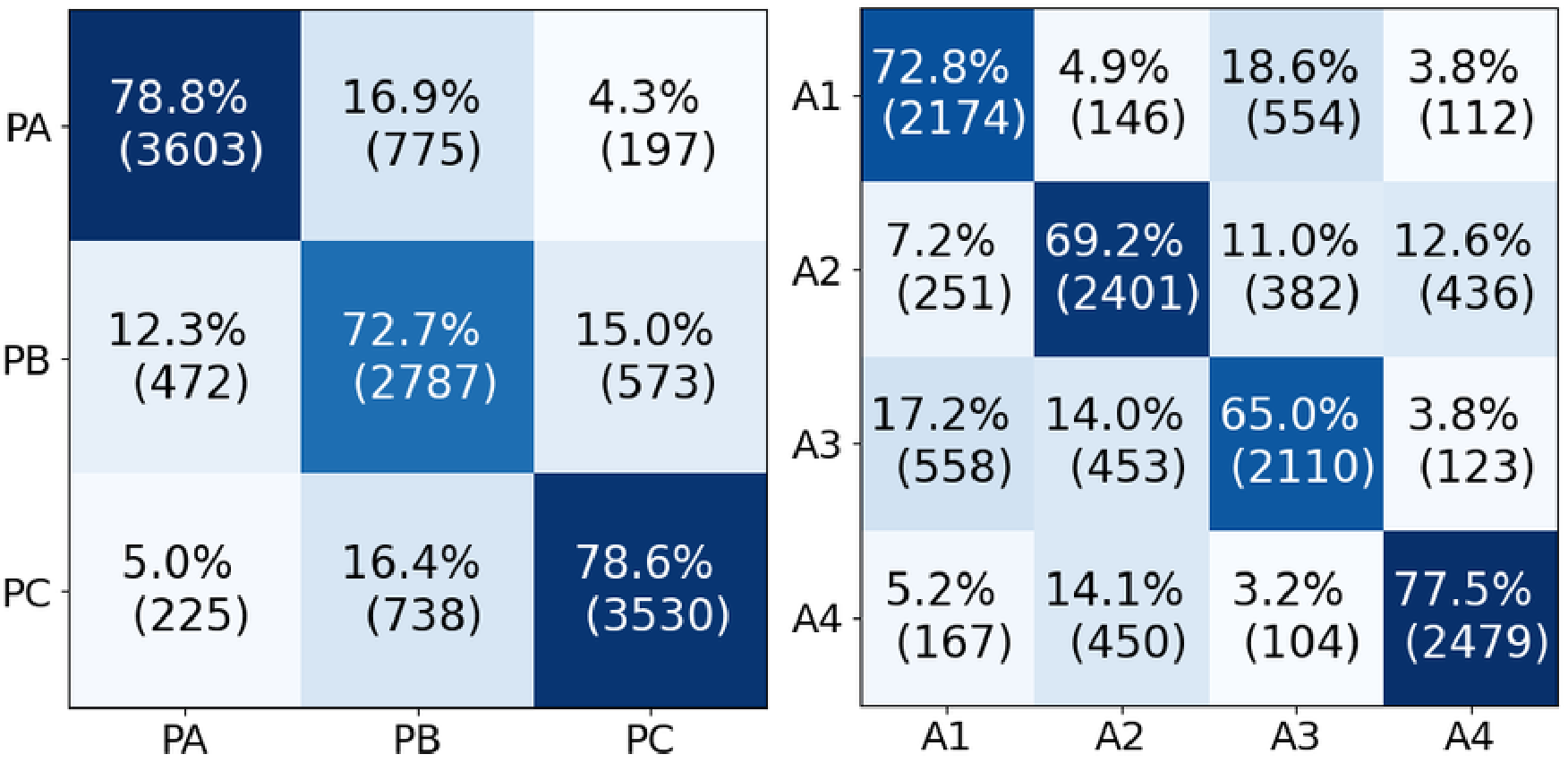}
  \end{center}
  \caption{Confusion matrices for three pose classification (left) and four class activity classification (right). The classification algorithm is XGB  with statistical features from radar. Each cell displays the percentage of correctly classified instances along with the actual number of windows.}
  \label{fig:clas_stat_pose_and_act}
\end{figure}

\begin{figure}[h!]
  \begin{center}
    \includegraphics*[width=1\columnwidth]{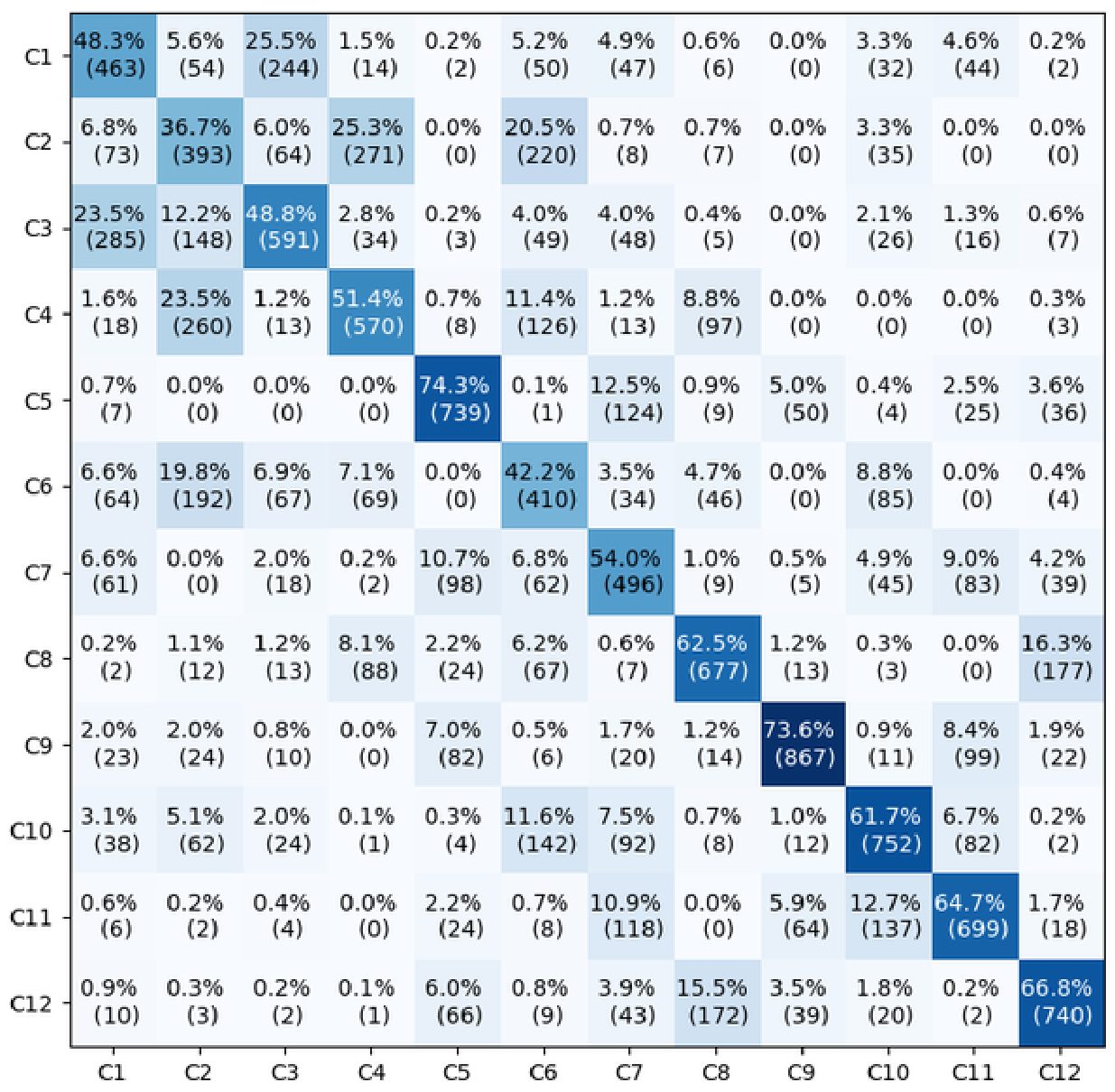}
  \end{center}
  \caption{Confusion matrix with XGB algorithm trained with all statistical features extracted from the radar. Each cell displays the percentage of of correctly classified instances along with the actual number of windows.}
  \label{fig:clas_stat_posAct}
\end{figure}

In this context, in Fig. \ref{fig:clas_stat_pose_and_act} we provide the results of the confusion matrices for pose identification and activity estimation separately using data from the radar with a XGB classifier. Pose estimation obtains 79\% accuracy for standing pose, 73\% for sitting pose and 79\% in lying down pose using extended statistical features. Breathing pattern estimation obtains 73\% accuracy for normal breathing, 69\% for reading, 65\% for guided breathing and 78\% for apnea showing the best result in this particular scenario. A more detailed analysis of 12-class is shown in Fig \ref{fig:clas_stat_posAct}.

\subsection{Physiological features and stratification}
For this experiment, we stratify both radar and camera data based on pose and breathing patterns. The pose classification models are trained and tested with only one specific breathing pattern at a time. Similarly, the breathing pattern classification experiment follows the same approach, where we train and test with only one specific pose at each iteration. For comparative purposes, we present results for pose classifications trained with all breathing patterns and breathing classifications trained with all poses concurrently. We use XGB as the machine learning classifier, incorporating both statistical and physiological features. The results are presented in Table \ref{tab:clas_fine}.

\begin{table*}[ht!]
\renewcommand{\arraystretch}{1.02}
\setlength{\tabcolsep}{1.3em}
\caption{Stratified classification with physiological features (XGB)}
\label{tab:clas_fine}
\centering
\scalebox{1.0}{
\begin{tabular}{lllll|llll}
 \multicolumn{1}{c}{} & \multicolumn{4}{c}{Radar features (184)} & \multicolumn{4}{c}{Camera features (308)} \\
Task &Accuracy &Precision &Recall &F1  &Accuracy &Precision &Recall &F1 \\
\hline
Pose results trained on: &&&&&&&&\\
Normal Breathing  &0.85 &0.85 &0.86 &0.83  &0.81 &0.80 &0.80 &0.78 \\
Reading           &0.79 &0.80 &0.82 &0.78  &0.76 &0.76 &0.78 &0.72\\                  
Guided Breathing  &0.67 &0.68 &0.70 &0.64  &0.68 &0.67 &0.67 &0.63\\                                 
Apnea             &0.90 &0.90 &0.91 &0.88  &0.84 &0.85 &0.84 &0.81\\    
All patterns &\textbf{0.77} &\textbf{0.77} &\textbf{0.79} &\textbf{0.76} &\textbf{0.77} &\textbf{0.76} &\textbf{0.78} &\textbf{0.75} \\
\hline
Breathing pattern results trained on:&&&&&&&&\\
Pose A            &0.61 &0.61 &0.61 &0.57   &0.66 &0.67 &0.70 &0.63\\ 
Pose B            &0.80 &0.80 &0.82 &0.78   &0.79 &0.79 &0.81 &0.76 \\
Pose C            &0.85 &0.85 &0.87 &0.83   &0.59 &0.58 &0.62 &0.54\\ 
All poses &\textbf{0.72} &\textbf{0.72} &\textbf{0.75} &\textbf{0.71} &\textbf{0.70} &\textbf{0.71} &\textbf{0.74} &\textbf{0.69} \\
\hline
Sex trained on:&&&&&&&&\\
All windows &\textbf{0.54} &\textbf{0.27} &\textbf{0.50} &\textbf{0.34} &\textbf{0.53} &\textbf{0.26} &\textbf{0.50} &\textbf{0.34} \\
 \hline
posAct trained on:&&&&&&&&\\
All windows  &\textbf{0.58} &\textbf{0.58} &\textbf{0.60} &\textbf{0.55}  &\textbf{0.53} &\textbf{0.52} &\textbf{0.53}&\textbf{0.48} \\
\end{tabular}}
\end{table*}

The performance of both pose and breathing pattern classification is slightly improved with the addition of physiological signals, but not in a very significant manner. However, when evaluating the classification of Pose and Breathing Pattern when training only on specific sessions, interesting observations emerge. In Pose estimation, models trained on apnea data demonstrate significantly better results at 90\% accuracy compared to other breathing patterns. We hypothesize that the steady pose during apnea eliminates subtle radar wave interferences, capturing only chest displacement information. Similar patterns are observed with models trained using camera features. When breathing pattern recognition tasks are studied separately for poses A, B, and C, Pose C (lying down) exhibits the highest accuracy at 85\%, followed by Pose B at 80\%, and Pose A at 61\%. 

\subsection{Radar and camera joint modalities}
We perform results by joining features for both modalities, camera and radar. We apply the fusion at the feature level (pre-fusion). Results for the XGB classifier are shown in Table \ref{tab:clas_joint}.
All classification results show a slight improvement compared with their unimodal counterparts. Specifically, Pose classification achieves an accuracy of 87\%, Breathing pattern classification 83\%, and \textit{posAct} 72\%, while sex classification still shows a performance only mildly better than random guess. 
Additionally, we present results for regression tasks and show them in Table \ref{tab:reg_joint}. The results improve over simple configurations, but not in a very significant manner.

\begin{table}[h!]
\renewcommand{\arraystretch}{1.1}
\setlength{\tabcolsep}{1.15em}
\caption{Classification with joint features from camera and radar (XGB)}
\label{tab:clas_joint}
\centering

\begin{tabular}{lcccc}
Task (477 features)  &Accuracy  &Precision  &Recall &F1\\
\hline
Pose      &0.87      &0.87       &0.89   &0.86 \\
Activity  &0.83      &0.83       &0.85   &0.82 \\
Sex       &0.52      &0.26       &0.50   &0.33 \\
posAct    &0.72      &0.72       &0.73   &0.69 \\
\end{tabular}
\end{table}

\begin{table}[h!]
\renewcommand{\arraystretch}{1.1}
\setlength{\tabcolsep}{1.9em}
\caption{Regression with joint features from camera and radar (SVR)}
\label{tab:reg_joint}
\centering

\begin{tabular}{lccc}
Task (477 features) & MAE  & RMSE  & MAPE  \\
\hline
Age           &4.61  &4.61  &14.22 \\
Height        &7.94  &7.94  &4.65  \\
Weight        &11.44 &11.44 &16.99 \\ 
\end{tabular}
\vspace{-3mm}
\end{table}

\subsection{Alternative validation scheme using joint features}
As an alternative validation, we perform the classification experiments on all fused modalities (camera and radar), using both a 5-fold validation scheme and a train/validation/test data split. Results for XGB classifier are shown in Table \ref{tab:clas_alternative}.

\begin{table}[h!]
\renewcommand{\arraystretch}{1.1}
\setlength{\tabcolsep}{1.15em}
\caption{Alternative validation classifiers with joint features}
\label{tab:clas_alternative}
\centering
\begin{tabular}{lcccc}
\multicolumn{5}{l}{Classifier with k-fold cross-validation (XGB)}\\
Task (477 features)  &Accuracy  &Precision  &Recall &F1\\
\hline
 Pose      &0.86  &0.86   &0.86  &0.86 \\
Activity  &0.81  &0.81   &0.81  &0.81 \\
Sex       &0.51  &0.51   &0.51  &0.51 \\
posAct    &0.71  &0.71   &0.72  &0.71 \\
\hline    
\multicolumn{5}{l}{Classifier with train-validation-test split approach (XGB)}\\
Task (477 features)  &Accuracy  &Precision  &Recall &F1\\
\hline
Pose      &0.89  &0.88   &0.89  &0.88 \\
Activity  &0.83  &0.83   &0.84  &0.83 \\
Sex       &0.49  &0.49   &0.49  &0.48 \\
posAct    &0.75  &0.74   &0.75  &0.74 \\   
\end{tabular}
\end{table}

The results are still very similar to those using LOSO validation, although those using data split are marginally better, most likely due to a validation set including several users. These results are mostly useful for comparative purposes, particularly in cases where training 50 separate models is considered impractical.

\section{Discussion and Analysis}
Our data collection approach aims to ensure that sensors primarily capture fundamental pose differences while minimizing the influence of external noise or individual physiological variations such as height.
Data collection was conducted under static conditions, with subjects remaining immobile or minimizing movement as much as possible. Guided exercises were conducted under standardized conditions, ensuring consistency in illumination, distance, and angle to the devices.

The chosen poses represent typical static positions individuals adopt at home, while the selected breathing activities aim to maximize signal differences between sensors in a calm home environment. Consequently, our dataset excludes physical activities involving considerable movement or reflecting physiological states of fatigue or exhaustion.
However, a limitation of the dataset is that emergency situations are simulated, thus leading to the possibility of introducing some bias. 

The specific conditions regarding the pose, angle, and positioning of the device during data capture are unlikely to be replicated in everyday scenarios due to the free movement of individuals within a household, resulting in varying distances and angles relative to the capturing devices. Models trained on this dataset to estimate pose or physiological indicators may exhibit poor generalization in environments where sensor positions differ.

Our protocol is annotated (providing labels and timestamps for pose and breathing pattern activities), standardized (ensuring uniform illumination, adjusting radar height to match individual height to avoid capturing individual particularities), and conducted in a static manner (users remain stationary to minimize sensor noise due to motion), enabling supervised capture of camera and radar response differences.
We argue that this approach is conducive to future fine-tuning using contrastive learning on new, unsupervised captures.

The classification of positional parameters such as pose and breathing pattern activity demonstrates good results. This indicates that data extracted from radar and camera contains adequate information to accurately distinguish between standing, sitting, and lying positions, as well as among the four breathing patterns used in this study.
This suggests that correctly identifying breathing patterns in stationary poses particularly depicting emergency situations can be achieved using RGB-D cameras and radars in a home scenario.

The results obtained for biometric characteristics analysis, conversely, indicate worse performance, suggesting that features extracted from radar and camera do not provide sufficient information about physiological parameters.
Specifically, the results for age, weight, and height closely resemble those of a naive regressor, while the results for sex classification closely resemble random guessing.
This aligns with other studies \cite{li2024realfacesrppg} that demonstrate the difficulty in extracting physiological and personal data using solely signal analysis.
Nevertheless, this provides an opportunity for future research that could focus on biometric data using different preprocessing and architectures.

\section{Conclusion}

We introduced OMuSense-23, a comprehensive multimodal dataset designed for breathing pattern recognition and physiological data measurement aiming to enhance health monitoring within a domestic environment.
To the best of our knowledge, this multimodal dataset is the first to offer biosignal data captured from three distinct sensors: a radio frequency-based mmWave radar, an RGB camera, and a depth camera encompassing 50 individuals with 50\% balanced sex distribution of participants.
The participants engaged in a sequence of three poses: standing, sitting, and lying, during which they performed four activities: normal breathing, reading, guided breathing patterns, and breath-holding to simulate apnea. These activities were conducted within a controlled environment where each action was precisely timed and labelled.
We provided a comprehensive description of the data collection protocol and methodologies, with the aim of facilitating easy replication of our dataset across diverse populations and conditions.

Furthermore, we presented preliminary analyses and baseline comparisons, evaluating models based on sensor types (camera, radar, and their combination) and a variety of features used in model training (statistical and physiological features).
Subsequently, we provided a fine-grained analysis targeting physiological data, where models were trained only with specific poses of breathing patterns.
Additionally, we offered a 12-class classifier delineating all possible combinations of pose and breathing activity estimation.
Lastly, we provided three different evaluation protocols of the dataset for future researchers (LOSO, k-fold cross-validation and train-validation-test data split).
These protocols aim to support future evaluations employing diverse approaches, including deep learning methodologies.
Our benchmark results, attained an accuracy of 87\% for three pose identification, 83\% for estimating four-class breathing patterns, and 72\% for combined 12-class pose and activity recognition.
Our regression analysis for physiological task estimation provided an MAE of 4.61 for age, 7.94 for height, and 11.44 for weight.

More work can be done in the future to address these challenges of extracting different biosignals from the raw data.
Future endeavors will concentrate on capturing data in uncontrolled environments, where the learned representations from OMuSense-23 could be leveraged.
Additionally, investigating data fusion methods between signals from different sensors could potentially enhance biometric and physiological characteristics measurements. Our goal is to refine models capable of detecting emergency situations in scenarios not encountered within the boundaries of OMuSense-23.
Finally, we have made the dataset publicly available to serve as a valuable resource for fellow researchers.

\section{References}
\bibliographystyle{IEEEbib}
\bibliography{references}

\end{document}